\documentclass[letterpaper]{sig-alternate-2013}

\permission{\copyright 2017 International World Wide Web Conference Committee (IW3C2), 
published under Creative Commons CC BY 4.0 License.}
\conferenceinfo{WWW 2017 Companion,}{April 3-7, 2017, Perth, Australia.}
\copyrightetc{ACM \the\acmcopyr}
\crdata{978-1-4503-4914-7/17/04. \\
http://dx.doi.org/10.1145/3041021.3054223 \\
\includegraphics{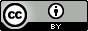}
}

\usepackage{multirow}
\usepackage{rotating}
\usepackage{booktabs}
\renewcommand\textfloatsep{2pt}
\renewcommand\floatsep{2pt}
\renewcommand\intextsep{2pt}
\renewcommand\dbltextfloatsep{2pt}
\renewcommand\dblfloatsep{2pt}
\usepackage[font=scriptsize,labelfont=bf]{caption}

\makeatletter
\def\@maketitle{\newpage
	\null
	\setbox\@acmtitlebox\vbox{%
		\baselineskip 20pt
		\vskip 2em                   
		\begin{center}
			{\ttlfnt \@title\par}       
			\vskip 1.5em                
			{\subttlfnt \the\subtitletext\par}\vskip 1.25em
			{\baselineskip 16pt\aufnt   
				\lineskip .5em             
				\begin{tabular}[t]{c}\@author
				\end{tabular}\par}
			\vskip 1.5em               
		\end{center}}
		\dimen0=\ht\@acmtitlebox
		\advance\dimen0 by -11.75pc\relax 
		\unvbox\@acmtitlebox
		\ifdim\dimen0<0.0pt\relax\vskip-\dimen0\fi}
\makeatother

\begin{document}
\title{Deep Learning for Hate Speech Detection in Tweets}

\numberofauthors{4}
\author{
Pinkesh Badjatiya$^1$\footnotemark[2] , Shashank Gupta$^1$\footnotemark[2] , Manish Gupta$^{1,2}$, Vasudeva Varma$^1$\\
\affaddr{
	$^1$IIIT-H, Hyderabad, India\ \ \ \ \ \ \ $^2$Microsoft, India
}\\
\email{\{pinkesh.badjatiya, shashank.gupta\}@research.iiit.ac.in, gmanish@microsoft.com, vv@iiit.ac.in}
}

\maketitle

\begin{abstract}
	Hate speech detection on Twitter is critical for applications like controversial event extraction, building AI chatterbots, content recommendation, and sentiment analysis. We define this task as being able to classify a tweet as racist, sexist or neither. The complexity of the natural language constructs makes this task very challenging. We perform extensive experiments with multiple deep learning architectures to learn semantic word embeddings to handle this complexity. Our experiments on a benchmark dataset of 16K annotated tweets show that such deep learning methods outperform state-of-the-art char/word n-gram methods by $\sim$18 F1 points.
\end{abstract}
{\renewcommand{\thefootnote}{\fnsymbol{footnote}}
	\footnotetext[2]{authors contributed equally}
}



\section{Introduction}
With the massive increase in social interactions on online social networks, there has also been an increase of hateful activities that exploit such infrastructure. On Twitter, hateful tweets are those that contain abusive speech targeting individuals (cyber-bullying, a politician, a celebrity, a product) or particular groups (a country, LGBT, a religion, gender, an organization, etc.). Detecting such hateful speech is important for analyzing public sentiment of a group of users towards another group, and for discouraging associated wrongful activities. It is also useful to filter tweets before content recommendation, or learning AI chatterbots from tweets\footnote{https://en.wikipedia.org/wiki/Tay\_(bot)}. 

The manual way of filtering out hateful tweets is not scalable, motivating researchers to identify automated ways. In this work, we focus on the problem of classifying a tweet as racist, sexist or neither. The task is quite challenging due to the inherent complexity of the natural language constructs -- different forms of hatred, different kinds of targets, different ways of representing the same meaning. Most of the earlier work revolves either around manual feature extraction~\cite{waseem2016hateful} or use representation learning methods followed by a linear classifier~\cite{djuric2015hate, nobata2016abusive}. However, recently deep learning methods have shown accuracy improvements across a large number of complex problems in speech, vision and text applications. To the best of our knowledge, we are the first to experiment with deep learning architectures for the hate speech detection task.


In this paper, we experiment with multiple classifiers such as Logistic Regression, Random Forest, SVMs, Gradient Boosted Decision Trees (GBDTs) and Deep Neural Networks(DNNs). The feature spaces for these classifiers are in turn defined by task-specific embeddings learned using three deep learning architectures: FastText, Convolutional Neural Networks (CNNs), Long Short-Term Memory Networks (LSTMs). As baselines, we compare with feature spaces comprising of char n-grams~\cite{waseem2016hateful}, TF-IDF vectors, and Bag of Words vectors (BoWV).

Main contributions of our paper are as follows: (1) We investigate the application of deep learning methods for the task of hate speech detection. (2) We explore various tweet semantic embeddings like char n-grams, word Term Frequency-Inverse Document Frequency (TF-IDF) values, Bag of Words Vectors (BoWV) over Global Vectors for Word Representation (GloVe), and task-specific embeddings learned using FastText, CNNs and LSTMs. (3) Our methods beat state-of-the-art methods by a large margin ($\sim$18 F1 points better).

\section{Proposed Approach}
We first discuss a few baseline methods and then discuss the proposed approach. In all these methods, an embedding is generated for a tweet and is used as its feature representation with a classifier.

\noindent\underline{\textbf{Baseline Methods}}: As baselines, we experiment with three broad representations. (1) Char n-grams: It is the state-of-the-art method~\cite{waseem2016hateful} which uses character n-grams for hate speech detection. (2) TF-IDF: TF-IDF are typical features used for text classification. (3) BoWV: Bag of Words Vector approach uses the average of the word (GloVe) embeddings to represent a sentence. We experiment with multiple classifiers for both the TF-IDF and the BoWV approaches.

\noindent\underline{\textbf{Proposed Methods}}: We investigate three neural network architectures for the task, described as follows. For each of the three methods, we initialize the word embeddings with either random embeddings or GloVe embeddings. (1) CNN: Inspired by Kim et. al~\cite{Kim14}'s work on using CNNs for sentiment classification, we leverage CNNs for hate speech detection. We use the same settings for the CNN as described in~\cite{Kim14}. (2) LSTM: Unlike feed-forward neural networks, recurrent neural networks like LSTMs can use their internal memory to process arbitrary sequences of inputs. Hence, we use LSTMs to capture long range dependencies in tweets, which may play a role in hate speech detection. (3) FastText: FastText~\cite{joulin2016bag} represents a document by average of word vectors similar to the BoWV model, but allows update of word vectors through Back-propagation during training as opposed to the static word representation in the BoWV model, allowing the model to fine-tune the word representations according to the task.

All of these networks are trained (fine-tuned) using labeled data with back-propagation. Once the network is learned, a new tweet is tested against the network which classifies it as racist, sexist or neither. Besides learning the network weights, these methods also learn task-specific word embeddings tuned towards the hate speech labels. Therefore, for each of the networks, we also experiment by using these embeddings as features and various other classifiers like SVMs and GBDTs as the learning method.

\section{Experiments}

\subsection{Dataset and Experimental Settings}
We experimented with a dataset of 16K annotated tweets made available by the authors of~\cite{waseem2016hateful}. Of the 16K tweets, 3383 are labeled as sexist, 1972 as racist, and the remaining are marked as neither sexist nor racist. For the embedding based methods, we used the GloVe~\cite{pennington2014glove} pre-trained word embeddings. GloVe embeddings\footnote{http://nlp.stanford.edu/projects/glove/} have been trained on a large tweet corpus (2B tweets, 27B tokens, 1.2M vocab, uncased). We experimented with multiple word embedding sizes for our task. We observed similar results with different sizes, and hence due to lack of space we report results using embedding size=200. We performed 10-Fold Cross Validation and calculated weighted macro precision, recall and F1-scores.

We use `adam' for CNN and LSTM, and `RMS-Prop' for FastText as our optimizer. We perform training in batches of size 128 for CNN \& LSTM and 64 for FastText. More details on the experimental setup can be found from our publicly available source code\footnote{https://github.com/pinkeshbadjatiya/twitter-hatespeech}.

\setlength{\tabcolsep}{0.15pc}
\begin{table}
	\scriptsize
	\caption{\scriptsize Comparison of Various Methods (Embedding Size=200 for GloVe as well as for Random Embedding)}
	\vspace{-0.6pc}
	\label{tab:experiments}
	\begin{tabular}{|p{1.05cm}|p{4.9cm}|l|p{0.7cm}|p{0.7cm}|}
		\hline
		&Method & Prec & Recall& F1\\
		\hline
		\multirow{5}{*}{\parbox{1.05cm}{\textbf{Part A}:\\ Baselines}}  &  Char n-gram+Logistic Regression~\cite{waseem2016hateful} & 0.729 & 0.778 & 0.753\\
		&		TF-IDF+Balanced SVM &	0.816 	&0.816 &0.816\\
		&		TF-IDF+GBDT &0.819&	0.807&	0.813\\
		&   BoWV+Balanced SVM& 0.791 & 0.788 & 0.789\\
		&		BoWV+GBDT& 0.800 & 0.802 & 0.801\\
		\hline
		\multirow{6}{*}{\parbox{1.05cm}{\textbf{Part B}:\\ DNNs Only}}  &    CNN+Random Embedding& 0.813& 0.816&0.814\\
		&    CNN+GloVe&0.839&0.840&0.839\\
		&    FastText+Random Embedding& 0.824& 0.827 & 0.825 \\
		&    FastText+GloVe& 0.828 & 0.831& 0.829\\
		&    LSTM+Random Embedding& 0.805& 0.804& 0.804\\
		&    LSTM+GLoVe& 0.807 & 0.809& 0.808\\
		\hline
		\multirow{6}{*}{\parbox{1.05cm}{\textbf{Part C}:\\ DNNs + GBDT Classifier}}  &		CNN+GloVe+GBDT&	0.864	&0.864&	0.864\\
		&CNN+Random Embedding+GBDT&0.864&0.864&0.864\\
		&FastText+GloVe+GBDT &0.853&0.854&0.853\\
		&FastText+Random Embedding+GBDT &0.886&0.887&0.886\\ 
		&LSTM+GloVe+GBDT& 0.849&0.848&0.848\\
		&LSTM+Random Embedding+GBDT&	\textbf{0.930}	&\textbf{0.930}&	\textbf{0.930}\\
		\hline
	\end{tabular}
\end{table}

\subsection{Results and Analysis}

Table~\ref{tab:experiments} shows the results of various methods on the hate speech detection task. Part A shows results for baseline methods. Parts B and C focus on the proposed methods where part B contains methods using neural networks only, while part C uses average of word embeddings learned by DNNs as features for GBDTs. We experimented with multiple classifiers but report results mostly for GBDTs only, due to lack of space. 

As the table shows, our proposed methods in part B are significantly better than the baseline methods in part A. Among the baseline methods, the word TF-IDF method is better than the character n-gram method. Among part B methods, CNN performed better than LSTM which was better than FastText. Surprisingly, initialization with random embeddings is slightly better than initialization with GloVe embeddings when used along with GBDT. Finally, part C methods are better than part B methods. The best method is ``LSTM + Random Embedding + GBDT'' where tweet embeddings were initialized to random vectors, LSTM was trained using back-propagation, and then learned embeddings were used to train a GBDT classifier. Combinations of CNN, LSTM, FastText embeddings as features for GBDTs did not lead to better results. Also note that the standard deviation for all these methods varies from 0.01 to 0.025. 

To verify the task-specific nature of the embeddings, we show top few similar words for a few chosen words in Table~\ref{tab:caseStudies} using the original GloVe embeddings and also embeddings learned using DNNs. The similar words obtained using deep neural network learned embeddings clearly show the ``hatred'' towards the target words, which is in general not visible at all in similar words obtained using GloVe.

\begin{table}%
	\centering
	\scriptsize
	\caption{\scriptsize Embeddings learned using DNNs clearly show the ``racist'' or ``sexist'' bias for various words.}
	\vspace{-0.6pc}
	\label{tab:caseStudies}
	\begin{tabular}{|p{2.5pc}|p{8pc}|p{8pc}|}
		\hline
		Target Word&Similar words using GloVe&Similar words using task-specific embeddings learned using DNNs \\
		\hline
		\hline
		pakistan&karachi, pakistani, lahore, india, taliban, punjab, islamabad&mohammed, murderer, pedophile, religion, terrorism, islamic, muslim\\
		\hline
		female&male, woman, females, women, girl, other, artist, girls, only, person&sexist, feminists, feminism, bitch, feminist, blonde, bitches, dumb, equality, models, cunt\\
		\hline
		muslims&christians, muslim, hindus, jews, terrorists, islam, sikhs, extremists, non-muslims, buddhists&islam, prophet, quran, slave, jews, slavery, pedophile, terrorist, terrorism, hamas, murder\\
		\hline
	\end{tabular}
	
\end{table} 

\section{Conclusions}
In this paper, we investigated the application of deep neural network architectures for the task of hate speech detection. We found them to significantly outperform the existing methods. Embeddings learned from deep neural network models when combined with gradient boosted decision trees led to best accuracy values. In the future, we plan to explore the importance of the user network features for the task.

\bibliographystyle{abbrv}
\scriptsize

\end{document}